\documentclass[letterpaper, 10 pt, conference]{ieeeconf} 

\IEEEoverridecommandlockouts
\PassOptionsToPackage{table}{xcolor} 
\usepackage{xcolor}
\usepackage{amsmath,amssymb} 
\usepackage{xcolor}          

\definecolor{cvprblue}{rgb}{0.21,0.49,0.74}
\definecolor{LightGreen}{rgb}{0.88,1,0.88}
\definecolor{LightRed}{rgb}{1,0.88,0.88}
\definecolor{BrightBrownBase}{rgb}{0.8,0.5,0.2}
\colorlet{BrightBrown}{BrightBrownBase!40!white}
\definecolor{verylightgray}{RGB}{245,245,245}

\usepackage{amsmath}
\usepackage{amssymb}
\usepackage{graphicx}
\usepackage{booktabs}
\usepackage{multirow}
\usepackage{array}
\usepackage{float}
\usepackage{tabularx}
\usepackage{colortbl}
\usepackage{makecell}
\usepackage{longtable}
\usepackage{adjustbox}
\usepackage{verbatim}
\usepackage{caption}  
\usepackage{fancyvrb}
\usepackage{ragged2e}
\usepackage{url}
\newcommand{\signtile}[3][1.2cm]{%
  \begin{minipage}[t]{\linewidth}
    \centering
    \includegraphics[height=#1,keepaspectratio]{#2}\par\vspace{2pt}%
    {\RaggedRight\footnotesize #3\par}%
  \end{minipage}%
}

\usepackage{subcaption}  
\usepackage{tikz}
\usetikzlibrary{positioning,shapes.geometric,arrows.meta,calc,patterns}
\usepackage{pgfplots}
\usepgfplotslibrary{polar,groupplots}
\usepackage{pgfplotstable}
\pgfplotsset{compat=1.16}

\title{\LARGE \bf
Distilling Vision-Language Models for Robust Traffic Sign Perception in Autonomous Vehicles
}

\author{
	\parbox{\textwidth}{%
		\centering
		Pedram MohajerAnsari, Amir Salarpour, Mert D. Pes\'{e}%
	}%
	\thanks{Pedram MohajerAnsari, Amir Salarpour, and Mert D. Pes\'{e} are with Clemson University, Clemson, SC, USA
		\{\texttt{pmohaje}, \texttt{asalarp}, \texttt{mpese}\}@clemson.edu}
}

\usepackage{graphicx}

\usepackage{booktabs}
\usepackage{multirow}
\usepackage{xcolor}
\usepackage{amssymb}  
\newcommand{\cmark}{{\color{green!70!black}\checkmark}}
\newcommand{\xmark}{{\color{red}$\boldsymbol{\times}$}}

\usepackage[hypcap=true]{caption}
\usepackage[hidelinks]{hyperref}
\hypersetup{draft}
\begin{document}

\begingroup
\setbox0=\hbox{%
  \includegraphics[width=1pt]{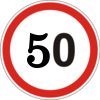}%
  \includegraphics[width=1pt]{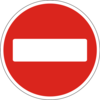}%
  \includegraphics[width=1pt]{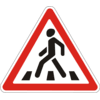}%
  \includegraphics[width=1pt]{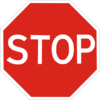}%
  \includegraphics[width=1pt]{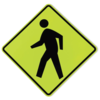}%
  \includegraphics[width=1pt]{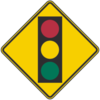}%
}
\endgroup

\maketitle
\thispagestyle{empty}
\pagestyle{empty}


\begin{abstract}
Traffic sign recognition (TSR) models based on deep neural networks
achieve strong clean-data performance but remain vulnerable to physically
realizable adversarial attacks, including shadow perturbations,
natural-light interference, and printed patches. Existing defenses often
improve robustness against one attack type while degrading performance on
others, and can reduce clean accuracy. We propose \textbf{LAMDA}
(\textbf{L}anguage-\textbf{A}nchored \textbf{M}odel for
\textbf{D}irection \textbf{A}lignment), a training framework that transfers
language-grounded structure into TSR models without using adversarial
examples or adding inference-time overhead. LAMDA builds two fixed
prototype banks from VLM-generated sign descriptions and class names using
a frozen OpenCLIP text encoder, and uses them to supervise visual features
through two complementary auxiliary losses during training. At inference,
the adapter and prototype banks are discarded, leaving a standard backbone
and classifier. Evaluated on GTSRB and LISA across four backbones and
three physical attack types, LAMDA is the only method among ten evaluated
that consistently improves robustness across all attack--backbone--dataset
combinations, with gains of up to $+12.5$\,pp under shadow attacks and
$+13.2$\,pp under natural-light attacks, while preserving or improving
clean accuracy in nearly all cases. Code is available at our repository:
\href{https://github.com/pedram-mohajer/LAMDA}
{\textcolor{blue}{\textit{\texttt{https://github.com/pedram-mohajer/LAMDA}}}}.
\end{abstract}

\section{Introduction}

Traffic sign recognition (TSR) is a fundamental component of modern autonomous systems, enabling vehicles to perceive and interpret traffic signs under diverse real-world conditions for safe driving operation~\cite{medina2025road}. Deep neural networks (DNNs) have become the standard solution for this task, with architectures such as ResNet~\cite{he2016deep}, Swin Transformer~\cite{liu2021swin}, and ViT~\cite{dosovitskiy2020image} achieving impressive accuracy on standard benchmarks. However, these models are vulnerable to adversarial examples (AEs), subtly modified inputs designed to fool DNNs~\cite{szegedy2013intriguing}, including physically realizable perturbations such as strategically placed shadows~\cite{zhong2022shadows}, natural-light interference~\cite{hsiao2024natural}, and printed patches placed directly on signs~\cite{eykholt2018robust}. Such attacks pose a critical threat to TSR, where a single misclassification can lead to unsafe driving decisions~\cite{vlmAV}.

Various defense methods have been proposed to mitigate such attacks:
defensive distillation reduces model sensitivity but struggles to
generalize across diverse perturbations~\cite{papernot2016distillation};
input transformations can suppress adversarial noise but often degrade
clean data quality, lowering accuracy~\cite{guo2017countering}; and
provable defenses, though theoretically robust, are computationally
expensive and difficult to scale~\cite{wong2018provable}.
Among these, adversarial training~\cite{madry2017towards} is the most
widely adopted defense, yet it requires adversarial examples at
training time and trades off clean accuracy for robustness~\cite{tsipras2018robustness,vlmAV}.


We introduce \textbf{LAMDA} (\textbf{L}anguage-\textbf{A}nchored
\textbf{M}odel for \textbf{D}irection \textbf{A}lignment), a training
framework that improves TSR robustness against physical adversarial
attacks without requiring any adversarial examples during training.
The core idea is to use the rich semantic structure of language as a
supervisory signal: before training begins, natural-language descriptions
of each sign class are generated using NVILA~\cite{liu2024nvila} and,
together with class names, are encoded by a frozen OpenCLIP text
encoder~\cite{cherti2023reproducible} to form two fixed prototype banks.
A lightweight adapter attached to the vision backbone projects image
features into the same space as these prototypes during training,
creating a channel through which language-grounded supervision
reaches the visual representation.
At inference, the adapter and prototype banks are discarded entirely,
leaving a standard backbone and classifier with no added overhead.

LAMDA trains the backbone with two complementary auxiliary losses
alongside standard cross-entropy.
The first pulls image features toward the language description of
the correct class, anchoring visual representations to the
appearance-level semantics of each sign.
The second uses the similarity between class names as soft training
targets for the classifier head, providing a richer supervisory
signal than one-hot labels.
The two losses target distinct aspects of robustness and are stronger
in combination than either is alone, trained entirely on clean data,
with no modifications to the inference pipeline.

We evaluate LAMDA on GTSRB~\cite{Stallkamp-IJCNN-2011} and a 16-class
LISA subset~\cite{GTDLBench_LISA_Traffic_Sign_website} across four
backbones (ResNet-18, ResNet-34, Swin-T, and ViT-B/16) trained
exclusively on clean data with no adversarial examples at any point
during training.
Adversarial examples are generated against two fixed target models,
\texttt{gtsrb-cnn} and \texttt{lisa-cnn}, under three physically
realizable attacks: shadow perturbations~\cite{zhong2022shadows},
natural-light interference~\cite{hsiao2024natural}, and the RP2
printable patch attack~\cite{eykholt2018robust}.
LAMDA is the only method that improves over the baseline on every evaluated attack, every backbone, and both datasets simultaneously, with gains of up to +12.5 pp under shadow attacks on GTSRB and +13.2 pp under natural-light attacks on LISA. Clean accuracy improves in seven of eight dataset–backbone settings, with only a minor 0.23 pp decrease for LISA ResNet-18.
In contrast, every one of the nine compared defenses degrades
performance on at least one combination of backbone, attack, and
dataset. The main contributions of this paper are:

\begin{itemize}

\item We propose LAMDA, a training framework that transfers
language-grounded robustness into TSR models by aligning visual
features with two fixed prototype banks, one built from
VLM-generated sign descriptions and one from class names,
through two complementary auxiliary losses.
LAMDA requires no adversarial examples during training, adds no overhead at inference, and is the only method among ten evaluated that consistently improves robustness across all evaluated attacks, backbones, and datasets, while improving clean accuracy in seven of eight dataset–backbone settings.

\item We conduct a systematic evaluation on GTSRB and LISA across
four backbones under three physically realizable attacks, comparing
LAMDA against nine re-implemented defenses under identical training
budgets.
A comprehensive ablation over the two loss weights confirms that the two losses are complementary, with the ($\lambda{=}1$, $\mu{=}1$) configuration providing the strongest overall performance across the evaluated conditions.

\end{itemize}
\section{Related Work}

Popp \emph{et al.} \cite{popp2024zero} propose distilling a compact image encoder using supervision from a large CLIP-style teacher while keeping the text side fixed and class representations available as precomputed guidance. A key takeaway for our setting is the ``text-off-device'' deployment view: training can leverage rich language supervision (including synthetic prompts/data), but inference can rely on a lightweight vision backbone. This aligns with our goal of transferring VLM-derived robustness cues into an efficient TSR model without running a text encoder on-board. Li \emph{et al.} \cite{li2024promptkd} distill vision-language knowledge using prompts while explicitly reusing pre-stored class text features as supervision. Their pipeline precomputes text embeddings once (per class/prompt) and trains the student primarily through logit/feature alignment against these frozen targets, enabling inference with a compact image encoder plus a fixed text-prototype bank. This is highly similar in spirit to our frozen name/description prototype supervision, differing mainly in our task-specific TSR formulation and prototype construction.

Wu \emph{et al.} \cite{wu2023tinyclip} introduce a CLIP distillation framework that compresses CLIP by mimicking cross-modal affinities (image--text similarity structure) and using weight inheritance to stabilize training. TinyCLIP is a strong baseline for ``CLIP-space'' distillation because it preserves the pairwise alignment geometry that underlies zero-shot prompting. In contrast, our approach targets an image-only TSR student at inference time and uses frozen text prototypes as supervision rather than retaining a full student text encoder. Yang \emph{et al.} \cite{yang2024clip} provide a systematic empirical study of CLIP distillation objectives, comparing feature matching, relational losses, contrastive objectives, and other KD variants across model scales. Their results clarify when simple feature/logit alignment is competitive and when additional relational constraints help, offering practical guidance for choosing robust and stable distillation signals. We leverage these insights to motivate our alignment losses against fixed language prototypes, while tailoring the objective to traffic-sign recognition and robustness.

\begin{table}[t]
\centering
\caption{Summary of baseline defenses compared against LAMDA, grouped by where they act in the pipeline. Each defense is applied on top of standard cross-entropy training under its best-performing hyperparameter configuration. \textbf{Note:} \textbf{\texttt{Adversarial training}}~\cite{madry2017towards} is excluded as all methods are evaluated under a clean-training-only protocol.}
\setlength{\tabcolsep}{3.5pt}
\renewcommand{\arraystretch}{1.1}
\scriptsize
\begin{tabular}{>{\centering\arraybackslash}m{0.12\linewidth} >{\centering\arraybackslash}m{0.18\linewidth} m{0.60\linewidth}}
\toprule
\textbf{Method} & \textbf{Category} & \textbf{Description} \\
\midrule
\textbf{LS}~\cite{szegedy2016rethinking}  & Train-time      & \emph{Label Smoothing}. Replaces one-hot labels with smoothed targets (e.g., $0.9$ for the correct class, $0.1/(C-1)$ otherwise) to reduce overconfidence. \\
\midrule
\textbf{GT}~\cite{bishop1995training}  & Train-time      & \emph{Gaussian Transformation}. Data augmentation with Gaussian blur/noise to encourage robustness to low-level distortions. \\
\midrule
\textbf{DO}~\cite{srivastava2014dropout}  & Train-time      & \emph{Dropout}. Randomly zeroes activations during training to reduce co-adaptation; acts as regularization. \\
\midrule
\textbf{JPG}~\cite{guo2017countering} & Preprocessing   & \emph{JPEG Compression}. Encodes/decodes images at fixed quality to suppress high-frequency artifacts (may remove useful detail). \\
\midrule
\textbf{BD}~\cite{xu2017feature}  & Preprocessing   & \emph{Bit Depth Reduction}. Quantizes intensities to fewer levels (e.g., 3--5 bits/channel), attenuating fine perturbations. \\
\midrule
\textbf{MED}~\cite{xu2017feature} & Preprocessing   & \emph{Median Filtering}. Applies a $3{\times}3$ median filter to smooth pixel noise, often blurring edges. \\
\midrule
\textbf{HEQ}~\cite{zuiderveld1994contrast}  & Preprocessing   & \emph{Histogram Equalization}. Adjusts intensity distributions to enhance contrast; can alter colors. \\
\midrule
\textbf{RRP}~\cite{xie2017mitigating} & Evaluation-time & \emph{Random Resize-and-Pad}. Randomly resizes the input and pads it to the original dimensions at test time, acting as a stochastic input transform. \\
\midrule
\textbf{RSE}~\cite{cohen2019certified} & Evaluation-time & \emph{Randomized Smoothing Ensemble}. Aggregates predictions over multiple noisy copies of the input to approximate a smoothed classifier. \\
\bottomrule
\end{tabular}
\vspace{-15pt}
\label{tab:baseline-defenses}
\end{table}

Dong \emph{et al.} \cite{dong2025robustifying} improve zero-shot robustness of VLMs by aligning representation subspaces induced by image augmentations and prompt variations (e.g., synonyms), and further aligning adversarial and clean subspaces via robust fine-tuning. This supports the broader thesis that language structure can act as an anchor for robust visual representations. Unlike their goal of robustifying the VLM itself for inference, we treat language features as frozen teachers and transfer robustness-relevant alignment into a compact TSR model that does not require a VLM at deployment.

\begin{table}[!t]
\centering
\caption{Representative GTSRB/LISA sign images used to produce image embeddings via the backbone$\rightarrow$adapter. Text prototypes ($E_{\text{desc}}$, $E_{\text{name}}$) are computed once with a frozen text encoder and kept fixed.}
\setlength{\tabcolsep}{2pt}
\renewcommand{\arraystretch}{0.4}
\scriptsize
\begin{tabularx}{\columnwidth}{p{0.08\columnwidth} *{3}{>{\RaggedRight\arraybackslash}X}}
\toprule
\textbf{Dataset} & \multicolumn{3}{c}{\textbf{Examples}} \\
\midrule
\textbf{GTSRB} &
\signtile[1.0cm]{Figures/Signs/gtsrb_speed_limit_50.png}{White sign, red border, number 50; max speed 50\,km/h.} &
\signtile[1.0cm]{Figures/Signs/gtsrb_no_entry.png}{Red sign with white bar; no entry.} &
\signtile[1.0cm]{Figures/Signs/gtsrb_pedestrians.png}{Triangular sign, red border; pedestrian crossing.} \\
\midrule
\textbf{LISA} &
\signtile[1.0cm]{Figures/Signs/list_stop_sing.png}{Octagonal red sign; mandatory stop.} &
\signtile[1.0cm]{Figures/Signs/lisa_pedestrian_crossing.png}{Yellow-green diamond; pedestrian crossing.} &
\signtile[1.0cm]{Figures/Signs/lisa_signal_ahead.png}{Yellow diamond; signal ahead.} \\
\bottomrule
\end{tabularx}
\vspace{-15pt}
\label{tab:proto-text-examples}
\end{table}

\section{LAMDA Framework}
\label{sec:3-defense_design}

Motivated by the strong robustness we observe from VLMs on \emph{unseen}
adversarial inputs, our goal is to transfer part of this robustness into
compact DNNs that can run in real time.
To this end, we introduce \textbf{LAMDA}
(Language-Anchored Model for Direction Alignment),
a VLM-inspired training method that uses text prototypes to guide a
standard vision backbone.
During training, these prototypes provide an extra signal that nudges
image features toward language-based directions.
At inference, the network is a standard vision backbone and classifier
with no VLM or text encoder in the loop,
so latency and memory remain comparable to a conventional AV model.

\begin{figure*}[t]
  \centering
  \includegraphics[width=0.9\textwidth]{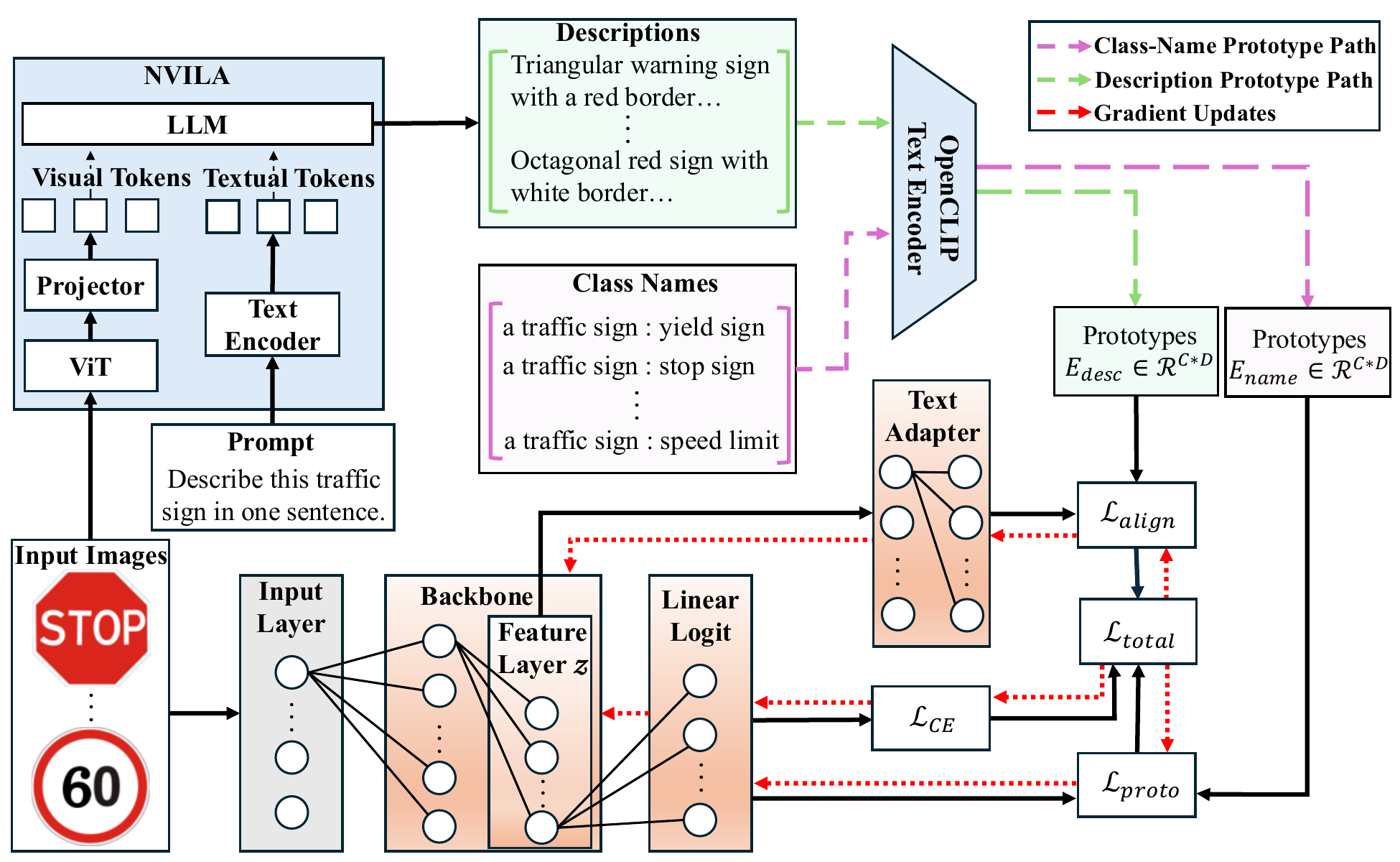}
  \caption{%
    \textbf{LAMDA (two-banks) training.}
    \textcolor{red!5!orange}{\textbf{Trainable}} nodes (backbone, linear
    head, adapter) are updated during training;
    \textcolor{blue!40}{\textbf{Frozen}} nodes are fixed
    (the OpenCLIP text encoder and prototype banks).
    A single frozen OpenCLIP text encoder is shared by both parallel
    branches: \emph{descriptions} are encoded once to build
    $\mathbf{E}_{\text{desc}}$ for adapter alignment via
    $\lambda\,\mathrm{CE}(\mathbf{S}_{\text{align}},y)$,
    where $\mathbf{S}_{\text{align}}=(\hat{\mathbf{t}}\,\mathbf{E}_{\text{desc}}^\top)/\tau$;
    \emph{class names} are encoded to $\mathbf{E}_{\text{name}}$,
    whose ground-truth row $\mathbf{t}_y$ induces soft targets
    $\sigma((\mathbf{t}_y\,\mathbf{E}_{\text{name}}^\top)/\tau_p)$
    for the head's prototype BCE weighted by $\mu$.
    Total loss:
    $\mathcal{L}=\mathrm{CE}
      +\lambda\,\mathrm{CE}_{\text{align}}
      +\mu\,\mathrm{BCE}_{\text{proto}}$.%
  }
  \label{fig:lamda_framework_twobanks}
\end{figure*}

LAMDA builds its text prototypes in two steps using off-the-shelf VLMs,
as illustrated in Figure~\ref{fig:lamda_framework_twobanks}.
First, for each traffic-sign image we query NVILA with the prompt:
\textit{``Describe the visual appearance of this traffic sign in one
sentence, including its shape, colors, border, background, and any
symbols, numbers, or text.''}
For each class we collect the resulting descriptions and also store a
short class name such as ``speed limit 50'' or ``no entry''.
Examples of image--description pairs are shown in
Table~\ref{tab:proto-text-examples}.

Second, we pass both the descriptions and the class names through a
frozen OpenCLIP text encoder to obtain vector embeddings.
We average and $\ell_2$-normalize these embeddings to form two fixed
prototype banks:
one built from descriptions ($\mathbf{E}_{\text{desc}}$) and one from
class names ($\mathbf{E}_{\text{name}}$).
Both banks are computed once before training and never updated. All descriptions and prototype banks are constructed exclusively from training-set images; no validation or test images are used.

\paragraph{Formal definition.}
Given an input $x$ with label $y\in\{1,\dots,C\}$,
define the backbone, head, and adapter mappings:
\[
  f_\theta:\;\mathbb{R}^{H\times W\times 3}\to\mathbb{R}^d,\quad
  h_W:\;\mathbb{R}^d\to\mathbb{R}^C,\quad
  g_\phi:\;\mathbb{R}^d\to\mathbb{R}^D.
\]
The backbone produces a visual feature vector and the head maps it to
class logits:
\[
  \mathbf{z} \;=\; f_\theta(x)\in\mathbb{R}^d,\qquad
  \mathbf{o} \;=\; h_W(\mathbf{z})\in\mathbb{R}^C.
\]
A lightweight adapter (two-layer MLP, hidden dim 512, batch norm, ReLU)
projects $\mathbf{z}$ into the shared text space and is $\ell_2$-normalized:
\[
  \hat{\mathbf{t}} \;=\; \mathrm{norm}\!\bigl(g_\phi(\mathbf{z})\bigr)\in\mathbb{R}^D.
\]

For each class $c$, descriptions are collected by prompting NVILA with
representative class images, then deduplicated and templated
(e.g., ``a traffic sign: \{description\}'').
Each prompt is encoded by the frozen text encoder, averaged, and
normalized to form $\mathbf{e}^{\text{desc}}_c \in \mathbb{R}^D$.
Likewise, class names are inserted into templates
(e.g., ``a traffic sign: \{name\}''), encoded, averaged, and normalized
to form $\mathbf{e}^{\text{name}}_c \in \mathbb{R}^D$.
Stacking across classes yields two fixed banks:
\[
  \mathbf{E}_{\text{desc}} \in \mathbb{R}^{C\times D},\qquad
  \mathbf{E}_{\text{name}} \in \mathbb{R}^{C\times D}.
\]
\begin{figure}[t]
\centering
\captionsetup[subfigure]{justification=centering}
\begin{subfigure}[t]{\linewidth}
  \centering
  \begin{minipage}[t]{0.44\linewidth}
    \centering
    \includegraphics[width=0.38\linewidth]{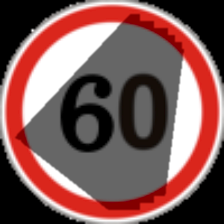}\hspace{0.02\linewidth}
    \includegraphics[width=0.38\linewidth]{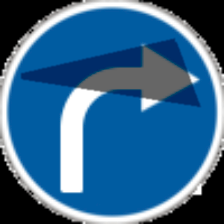}
    \par\vspace{1pt}\footnotesize\textbf{GTSRB}
  \end{minipage}\hfill
  \begin{minipage}[t]{0.44\linewidth}
    \centering
    \includegraphics[width=0.38\linewidth]{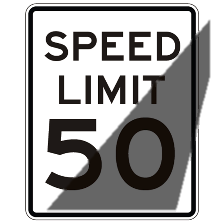}\hspace{0.02\linewidth}
    \includegraphics[width=0.38\linewidth]{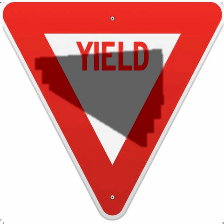}
    \par\vspace{1pt}\footnotesize\textbf{LISA}
  \end{minipage}
  \vspace{0.2em}
  \caption{Shadows~\cite{zhong2022shadows}}
\end{subfigure}
\vspace{0.3em}
\begin{subfigure}[t]{\linewidth}
  \centering
  \begin{minipage}[t]{0.44\linewidth}
    \centering
    \includegraphics[width=0.38\linewidth]{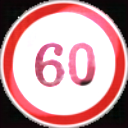}\hspace{0.02\linewidth}
    \includegraphics[width=0.38\linewidth]{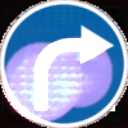}
    \par\vspace{1pt}\footnotesize\textbf{GTSRB}
  \end{minipage}\hfill
  \begin{minipage}[t]{0.44\linewidth}
    \centering
    \includegraphics[width=0.38\linewidth]{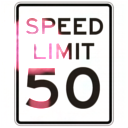}\hspace{0.02\linewidth}
    \includegraphics[width=0.38\linewidth]{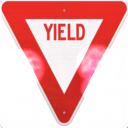}
    \par\vspace{1pt}\footnotesize\textbf{LISA}
  \end{minipage}
  \vspace{0.2em}
  \caption{Natural Light~\cite{hsiao2024natural}}
\end{subfigure}
\vspace{0.3em}
\begin{subfigure}[t]{\linewidth}
  \centering
  \begin{minipage}[t]{0.44\linewidth}
    \centering
    \includegraphics[width=0.38\linewidth]{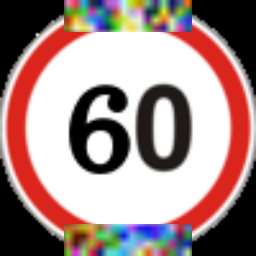}\hspace{0.02\linewidth}
    \includegraphics[width=0.38\linewidth]{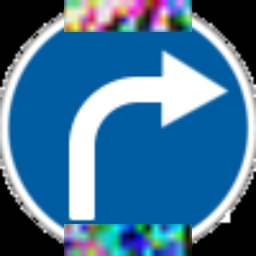}
    \par\vspace{1pt}\footnotesize\textbf{GTSRB}
  \end{minipage}\hfill
  \begin{minipage}[t]{0.44\linewidth}
    \centering
    \includegraphics[width=0.38\linewidth]{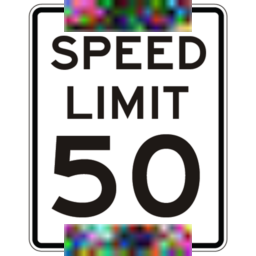}\hspace{0.02\linewidth}
    \includegraphics[width=0.38\linewidth]{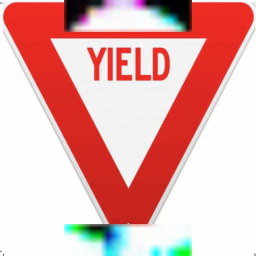}
    \par\vspace{1pt}\footnotesize\textbf{LISA}
  \end{minipage}
  \vspace{0.2em}
  \caption{RP2~\cite{eykholt2018robust}}
\end{subfigure}
\caption{Examples of physical perturbations. (a) \emph{Shadows}~\cite{zhong2022shadows}. (b) \emph{Natural Light}~\cite{hsiao2024natural}. (c) \emph{RP2}~\cite{eykholt2018robust}.}
\vspace{-15pt}
\label{fig:shadow-light-examples}
\end{figure}
\paragraph{Alignment loss.}
The adapter output is compared to description prototypes via
cosine-similarity logits with temperature $\tau>0$:
\[
  \mathbf{S}_{\text{align}}(x)
  \;=\; \frac{\hat{\mathbf{t}}\,\mathbf{E}_{\text{desc}}^\top}{\tau}
  \;\in\mathbb{R}^C.
\]
An auxiliary cross-entropy treats $\mathbf{S}_{\text{align}}$ as logits
for the target class, yielding the alignment loss:
\[
  \mathcal{L}_{\text{align}}(x,y)
  \;=\; \mathrm{CE}\!\bigl(\mathbf{S}_{\text{align}}(x),\,y\bigr).
\]
This term pulls $\hat{\mathbf{t}}$ toward the description prototype of
the target class and away from all others, anchoring backbone features
to semantic directions defined by natural language.
The weight $\lambda \geq 0$, linearly warmed up in early epochs,
controls the contribution of this term in the full objective.

\paragraph{Prototype loss.}
The name prototypes induce a second regulariser at the head.
For label $y$, retrieve its prototype
$\mathbf{t}_y = \mathbf{e}^{\text{name}}_y \in \mathbb{R}^D$
and form soft targets by comparing against the full name bank with
temperature $\tau_p > 0$ via the element-wise sigmoid $\sigma$:
\[
  \mathbf{q}
  \;=\; \sigma\!\bigl((\mathbf{t}_y\,\mathbf{E}_{\text{name}}^\top)/\tau_p\bigr)
  \;\in[0,1]^C.
\]
Note that $\sigma$ denotes the sigmoid (not softmax), so each entry of
$\mathbf{q}$ is an independent soft target in $[0,1]$, reflecting the
degree of semantic similarity between the target class and every other
class name.
These soft targets are used in a prototype-based binary cross-entropy,
where the head logits are scaled by a temperature $\tau_s > 0$ before
comparison:
\[
  \mathcal{L}_{\text{proto}}(x,y)
  \;=\; \mathrm{BCE}\!\bigl(\mathbf{o}/\tau_s,\;\mathbf{q}\bigr).
\]
The weight $\mu \geq 0$ controls this term's contribution in the full
objective.
This loss encourages the head logits to reflect the semantic
neighbourhood of the target class name, providing a smoother
supervisory signal than one-hot labels.

\paragraph{Full objective.}
The total training loss combines standard supervision with both
auxiliary terms, each weighted by their respective scalars:
\[
  \boxed{%
    \mathcal{L}
    \;=\;
    \mathcal{L}_{\text{CE}}
    \;+\;
    \lambda\,\mathcal{L}_{\text{align}}
    \;+\;
    \mu\,\mathcal{L}_{\text{proto}}
  }
\]
where $\mathcal{L}_{\text{CE}}=\mathrm{CE}(\mathbf{o},y)$ is the standard
classification loss on the head.
Together, the two banks create complementary pressures:
descriptions provide rich language grounding via
$\mathcal{L}_{\text{align}}$,
and class names enforce prototype-level regularisation via
$\mathcal{L}_{\text{proto}}$.
At inference, only the standard logits $\mathbf{o}$ are used for prediction;
the text banks are not needed.


\section{Evaluation of LAMDA}
\label{sec:4-evaluation}

We evaluate LAMDA on two public benchmarks for traffic-sign recognition:
GTSRB with $C{=}43$ classes and LISA, where we follow the common
$16$-class subset.
Training uses only the standard benign split for all methods;
no adversarial examples are included during training.
We study four backbones spanning CNNs and Transformers:
ResNet-18, ResNet-34, Swin-T, and ViT-B/16.
All methods share the same training schedule, augmentations, and data budget.

\begin{table*}[t]
\centering
\caption{GTSRB: Baseline accuracy and $\Delta$ accuracy (pp) vs.\ the
CE-only baseline. $\uparrow$/$\downarrow$ indicate improvement/degradation.
LAMDA ($\lambda{=}1$, $\mu{=}1$) is the only method that improves
over the baseline on all four splits across all backbones.}
\setlength{\tabcolsep}{3pt}
\renewcommand{\arraystretch}{1.6}
\footnotesize
\resizebox{\linewidth}{!}{%
\begin{tabular}{l|c|c|c|c|c|c|c|c|c|c|c|c|c|c|c|c}
\toprule
\multirow{2}{*}{\textbf{Method}} &
\multicolumn{4}{c|}{\textbf{ResNet-18}} &
\multicolumn{4}{c|}{\textbf{ResNet-34}} &
\multicolumn{4}{c|}{\textbf{Swin-T}} &
\multicolumn{4}{c}{\textbf{ViT-B/16}} \\
\cmidrule(lr){2-5}\cmidrule(lr){6-9}\cmidrule(lr){10-13}\cmidrule(lr){14-17}
& \textbf{benign} & \textbf{light} & \textbf{shadow} & \textbf{RP2}
& \textbf{benign} & \textbf{light} & \textbf{shadow} & \textbf{RP2}
& \textbf{benign} & \textbf{light} & \textbf{shadow} & \textbf{RP2}
& \textbf{benign} & \textbf{light} & \textbf{shadow} & \textbf{RP2} \\
\midrule
\rowcolor{gray!10}
\textbf{Baseline}
& 96.010 & 75.770 & 61.379 & 50.43
& 96.078 & 73.833 & 56.663 & 55.27
& 99.477 & 75.625 & 71.065 & 66.91
& 98.981 & 71.121 & 68.110 & 58.91 \\
\shortstack{$\lambda=1$ $\mu=1$}
& \cellcolor{green!20}{$\uparrow \mathbf{3.980}$} & \cellcolor{green!20}{$\uparrow \mathbf{3.641}$} & \cellcolor{green!20}{$\uparrow \mathbf{12.502}$} & \cellcolor{green!20}{$\uparrow$ \underline{4.351}}
& \cellcolor{green!20}{$\uparrow \mathbf{3.912}$} & \cellcolor{green!20}{$\uparrow \mathbf{2.267}$} & \cellcolor{green!20}{$\uparrow \mathbf{9.278}$} & \cellcolor{green!20}{$\uparrow$ \textbf{5.087}}
& \cellcolor{green!20}{$\uparrow \mathbf{0.513}$} & \cellcolor{green!20}{$\uparrow \mathbf{0.997}$} & \cellcolor{green!20}{$\uparrow \mathbf{3.322}$} & \cellcolor{green!20}{$\uparrow$ \underline{2.219}}
& \cellcolor{green!20}{$\uparrow \mathbf{1.010}$} & \cellcolor{green!20}{$\uparrow \underline{4.290}$} & \cellcolor{green!20}{$\uparrow \mathbf{8.231}$} & \cellcolor{green!20}{$\uparrow$ 1.893} \\
\textbf{LS}
& \cellcolor{green!20}{$\uparrow \underline{0.203}$} & \cellcolor{red!20}{$\downarrow 3.709$} & \cellcolor{red!20}{$\downarrow 3.709$} & \cellcolor{red!20}{$\downarrow 2.670$}
& \cellcolor{red!20}{$\downarrow 3.564$} & \cellcolor{red!20}{$\downarrow 4.881$} & \cellcolor{red!20}{$\downarrow 7.079$} & \cellcolor{red!20}{$\downarrow 0.520$}
& \cellcolor{red!20}{$\downarrow 2.469$} & \cellcolor{red!20}{$\downarrow 3.758$} & \cellcolor{red!20}{$\downarrow 11.710$} & \cellcolor{red!20}{$\downarrow 8.680$}
& \cellcolor{red!20}{$\downarrow 4.491$} & \cellcolor{red!20}{$\downarrow 11.011$} & \cellcolor{red!20}{$\downarrow 18.711$} & \cellcolor{red!20}{$\downarrow 4.190$} \\
\textbf{GT}
& \cellcolor{red!20}{$\downarrow 3.235$} & \cellcolor{red!20}{$\downarrow 7.118$} & \cellcolor{red!20}{$\downarrow 12.609$} & \cellcolor{red!20}{$\downarrow 6.070$}
& \cellcolor{red!20}{$\downarrow 4.503$} & \cellcolor{green!20}{$\uparrow \underline{1.801}$} & \cellcolor{red!20}{$\downarrow 13.006$} & \cellcolor{red!20}{$\downarrow 2.640$}
& \cellcolor{red!20}{$\downarrow 2.121$} & \cellcolor{green!20}{$\uparrow \underline{0.881}$} & \cellcolor{red!20}{$\downarrow 13.000$} & \cellcolor{red!20}{$\downarrow 0.930$}
& \cellcolor{red!20}{$\downarrow 6.505$} & \cellcolor{red!20}{$\downarrow 14.468$} & \cellcolor{red!20}{$\downarrow 23.301$} & \cellcolor{red!20}{$\downarrow 2.860$} \\
\textbf{DO}
& \cellcolor{red!20}{$\downarrow 2.169$} & \cellcolor{red!20}{$\downarrow 1.133$} & \cellcolor{red!20}{$\downarrow 8.261$} & \cellcolor{red!20}{$\downarrow 7.330$}
& \cellcolor{red!20}{$\downarrow 3.109$} & \cellcolor{red!20}{$\downarrow 0.010$} & \cellcolor{red!20}{$\downarrow 6.856$} & \cellcolor{red!20}{$\downarrow 0.960$}
& \cellcolor{red!20}{$\downarrow 1.850$} & \cellcolor{red!20}{$\downarrow 3.283$} & \cellcolor{red!20}{$\downarrow 10.064$} & \cellcolor{red!20}{$\downarrow 1.480$}
& \cellcolor{red!20}{$\downarrow 5.198$} & \cellcolor{red!20}{$\downarrow 18.245$} & \cellcolor{red!20}{$\downarrow 20.154$} & \cellcolor{red!20}{$\downarrow 1.950$} \\
\textbf{JPG}
& \cellcolor{red!20}{$\downarrow 1.811$} & \cellcolor{red!20}{$\downarrow 4.639$} & \cellcolor{red!20}{$\downarrow 11.786$} & \cellcolor{green!20}{$\uparrow \mathbf{8.740}$}
& \cellcolor{red!20}{$\downarrow 0.755$} & \cellcolor{red!20}{$\downarrow 2.266$} & \cellcolor{red!20}{$\downarrow 8.212$} & \cellcolor{green!20}{$\uparrow \underline{4.570}$}
& \cellcolor{red!20}{$\downarrow 0.736$} & \cellcolor{red!20}{$\downarrow 4.610$} & \cellcolor{red!20}{$\downarrow 8.030$} & \cellcolor{red!20}{$\downarrow 8.960$}
& \cellcolor{green!20}{$\uparrow \mathbf{0.119}$} & \cellcolor{red!20}{$\downarrow 2.315$} & \cellcolor{red!20}{$\downarrow 0.785$} & \cellcolor{red!20}{$\downarrow 0.770$} \\
\textbf{BD}
& \cellcolor{red!20}{$\downarrow 0.048$} & \cellcolor{red!20}{$\downarrow 0.174$} & \cellcolor{red!20}{$\downarrow 0.349$} & \cellcolor{red!20}{$\downarrow 4.720$}
& \cellcolor{red!20}{$\downarrow 0.058$} & \cellcolor{green!20}{$\uparrow \mathbf{0.029}$} & \cellcolor{red!20}{$\downarrow 0.232$} & \cellcolor{red!20}{$\downarrow 2.730$}
& \cellcolor{red!20}{$\downarrow 0.048$} & \cellcolor{red!20}{$\downarrow 0.320$} & \cellcolor{green!20}{$\uparrow \underline{2.274}$} & \cellcolor{green!20}{$\uparrow \mathbf{4.200}$}
& \cellcolor{green!20}{$\uparrow \underline{0.535}$} & \cellcolor{red!20}{$\downarrow 4.664$} & \cellcolor{red!20}{$\downarrow 2.873$} & \cellcolor{green!20}{$\uparrow \mathbf{9.030}$} \\
\textbf{MED}
& \cellcolor{red!20}{$\downarrow 4.823$} & \cellcolor{red!20}{$\downarrow 7.544$} & \cellcolor{red!20}{$\downarrow 12.841$} & \cellcolor{red!20}{$\downarrow 5.830$}
& \cellcolor{red!20}{$\downarrow 5.317$} & \cellcolor{red!20}{$\downarrow 7.292$} & \cellcolor{red!20}{$\downarrow 12.841$} & \cellcolor{red!20}{$\downarrow 9.350$}
& \cellcolor{red!20}{$\downarrow 3.409$} & \cellcolor{red!20}{$\downarrow 6.866$} & \cellcolor{red!20}{$\downarrow 9.745$} & \cellcolor{green!20}{$\uparrow \underline{1.053}$}
& \cellcolor{red!20}{$\downarrow 2.564$} & \cellcolor{red!20}{$\downarrow 2.469$} & \cellcolor{red!20}{$\downarrow 5.763$} & \cellcolor{green!20}{$\uparrow \underline{7.410}$} \\
\textbf{RRP}
& \cellcolor{red!20}{$\downarrow 22.003$} & \cellcolor{red!20}{$\downarrow 8.348$} & \cellcolor{red!20}{$\downarrow 29.421$} & \cellcolor{green!20}{$\uparrow 0.720$}
& \cellcolor{red!20}{$\downarrow 22.632$} & \cellcolor{red!20}{$\downarrow 12.251$} & \cellcolor{red!20}{$\downarrow 27.561$} & \cellcolor{red!20}{$\downarrow 0.910$}
& \cellcolor{red!20}{$\downarrow 4.222$} & \cellcolor{red!20}{$\downarrow 6.982$} & \cellcolor{red!20}{$\downarrow 14.529$} & \cellcolor{red!20}{$\downarrow 1.370$}
& \cellcolor{red!20}{$\downarrow 2.980$} & \cellcolor{red!20}{$\downarrow 0.349$} & \cellcolor{red!20}{$\downarrow 11.544$} & \cellcolor{red!20}{$\downarrow 2.190$} \\
\textbf{HEQ}
& \cellcolor{green!20}{$\uparrow 0.001$} & \cellcolor{red!20}{$\downarrow 0.001$} & \cellcolor{green!20}{$\uparrow \underline{0.001}$} & \cellcolor{red!20}{$\downarrow 0.050$}
& \cellcolor{red!20}{$\downarrow 0.001$} & \cellcolor{green!20}{$\uparrow 0.001$} & \cellcolor{red!20}{$\downarrow 0.001$} & \cellcolor{red!20}{$\downarrow 0.100$}
& \cellcolor{green!20}{$\uparrow \underline{0.001}$} & \cellcolor{red!20}{$\downarrow 0.001$} & \cellcolor{green!20}{$\uparrow \underline{3.000}$} & \cellcolor{red!20}{$\downarrow 0.020$}
& \cellcolor{green!20}{$\uparrow \underline{0.555}$} & \cellcolor{green!20}{$\uparrow \mathbf{5.249}$} & \cellcolor{green!20}{$\uparrow \underline{4.466}$} & \cellcolor{red!20}{$\downarrow 0.070$} \\
\textbf{RSE}
& \cellcolor{red!20}{$\downarrow 3.651$} & \cellcolor{red!20}{$\downarrow 6.847$} & \cellcolor{red!20}{$\downarrow 11.737$} & \cellcolor{green!20}{$\uparrow 0.430$}
& \cellcolor{red!20}{$\downarrow 0.668$} & \cellcolor{red!20}{$\downarrow 2.692$} & \cellcolor{red!20}{$\downarrow 5.472$} & \cellcolor{red!20}{$\downarrow 0.720$}
& \cellcolor{red!20}{$\downarrow 0.039$} & \cellcolor{red!20}{$\downarrow 2.896$} & \cellcolor{red!20}{$\downarrow 1.639$} & \cellcolor{red!20}{$\downarrow 6.040$}
& \cellcolor{red!20}{$\downarrow 0.484$} & \cellcolor{green!20}{$\uparrow \underline{1.651}$} & \cellcolor{green!20}{$\uparrow \underline{3.572}$} & \cellcolor{red!20}{$\downarrow 6.600$} \\
\bottomrule
\multicolumn{17}{l}{\footnotesize Per column, \textbf{bold} = largest improvement, \underline{underline} = second largest among improving (green) cells.}\\
\end{tabular}
}
\vspace{-15pt}
\label{tab:gtsrb-delta-compact}
\end{table*}

We evaluate against three physically realizable attacks,
representative examples of which are shown in
Figure~\ref{fig:shadow-light-examples}.
The shadow attack~\cite{zhong2022shadows} synthesizes shadows by applying
a polygonal mask over the sign region and attenuating the CIELAB
$\mathrm{L}$ channel; mask parameters are optimized as a black-box
attack using Expectation Over Transformations to preserve physical
realizability.
The natural-light attack~\cite{hsiao2024natural} applies illumination
via a parametric mask, modeling interference such as sunlight,
headlights, and reflections, with parameters searched by zeroth-order
optimization to maximize misclassification.
The RP2 attack~\cite{eykholt2018robust} optimizes a printable patch
constrained to the sign surface using a victim dataset captured under
varying viewing angles and distances, with synthetic transformations
to simulate physical conditions.
All adversarial examples are generated against fixed target models
and kept unchanged for evaluation; methods that involve training
are trained only on clean data and never see adversarial examples.

We compare LAMDA ($\lambda{=}1$, $\mu{=}1$) against nine widely used
defenses spanning three categories, summarized in
Table~\ref{tab:baseline-defenses}: train-time methods that modify
the learning signal or augmentations (label smoothing, Gaussian
transformation, dropout); preprocessing methods that apply a fixed
input transformation at inference (JPEG compression, bit-depth
reduction, median filtering, histogram equalization,
random resize-and-pad); and evaluation-time methods that change
prediction via test-time stochastic transforms without updating
parameters (randomized smoothing).
Each defense is integrated into or on top of standard cross-entropy
training ($\lambda{=}0$, $\mu{=}0$), which serves as the reference
baseline.

Tables~\ref{tab:gtsrb-delta-compact} and~\ref{tab:lisa-delta-compact}
report top-1 accuracy and $\Delta$ versus the baseline on GTSRB and
LISA respectively; for each defense, we sweep its key hyperparameters
and report the best-performing configuration.
LAMDA is the only method that consistently improves over the baseline
across adversarial splits, backbones, and datasets, while clean
accuracy is preserved or improved in nearly all settings.
(LISA ResNet-18: $-0.23$\,pp).
On GTSRB, gains are largest on shadow: $+12.50$\,pp (ResNet-18),
$+9.28$\,pp (ResNet-34), $+8.23$\,pp (ViT-B/16), and $+3.32$\,pp
(Swin-T).
RP2 and AE-light also improve consistently across all four backbones,
and benign accuracy is not degraded — it increases by $+3.98$\,pp and
$+3.91$\,pp for ResNet-18 and ResNet-34, and by $+1.01$\,pp and
$+0.51$\,pp for ViT-B/16 and Swin-T.
On LISA, the dominant split shifts to AE-light: $+13.20$\,pp
(ResNet-34) and $+11.30$\,pp (Swin-T), reflecting the lower
baseline accuracy on light attacks in LISA (42.3\% and 40.8\%
respectively) where the alignment signal provides the largest uplift.
Shadow and RP2 also improve consistently across all backbones
(shadow: $+10.15$\,pp ViT-B/16, $+6.45$\,pp ResNet-34;
RP2: $+7.89$\,pp ResNet-34, $+3.57$\,pp ResNet-18).

\begin{table*}[t]
\centering
\caption{LISA: Baseline accuracy and $\Delta$ accuracy (pp) vs.\ the
CE-only baseline. $\uparrow$/$\downarrow$ indicate improvement/degradation.
LAMDA ($\lambda{=}1$, $\mu{=}1$) is the only method that improves all three adversarial splits across all backbones, while improving benign accuracy for three of four backbones.}
\setlength{\tabcolsep}{3pt}
\renewcommand{\arraystretch}{1.6}
\footnotesize
\resizebox{\linewidth}{!}{%
\begin{tabular}{l|c|c|c|c|c|c|c|c|c|c|c|c|c|c|c|c} 
\toprule
\multirow{2}{*}{\textbf{Method}} &
\multicolumn{4}{c|}{\textbf{ResNet-18}} &
\multicolumn{4}{c|}{\textbf{ResNet-34}} &
\multicolumn{4}{c|}{\textbf{Swin-T}} &
\multicolumn{4}{c}{\textbf{ViT-B/16}} \\
\cmidrule(lr){2-5}\cmidrule(lr){6-9}\cmidrule(lr){10-13}\cmidrule(lr){14-17}
& \textbf{benign} & \textbf{light} & \textbf{shadow} & \textbf{RP2}
& \textbf{benign} & \textbf{light} & \textbf{shadow} & \textbf{RP2}
& \textbf{benign} & \textbf{light} & \textbf{shadow} & \textbf{RP2}
& \textbf{benign} & \textbf{light} & \textbf{shadow} & \textbf{RP2} \\
\midrule
\rowcolor{gray!10}
\textbf{Baseline}
& 99.342 & 48.412 & 61.586 & 51.482
& 98.561 & 42.273 & 58.360 & 56.751
& 98.707 & 40.789 & 60.543 & 53.002
& 98.854 & 43.261 & 60.172 & 43.604 \\
\shortstack{$\lambda=1$ $\mu=1$}
& \cellcolor{red!20}{$\downarrow 0.227$} & \cellcolor{green!20}{$\uparrow \mathbf{2.446}$} & \cellcolor{green!20}{$\uparrow \mathbf{4.704}$} & \cellcolor{green!20}{$\uparrow \mathbf{3.574}$}
& \cellcolor{green!20}{$\uparrow \mathbf{0.876}$} & \cellcolor{green!20}{$\uparrow \mathbf{13.202}$} & \cellcolor{green!20}{$\uparrow \mathbf{6.452}$} & \cellcolor{green!20}{$\uparrow \mathbf{7.890}$}
& \cellcolor{green!20}{$\uparrow \underline{0.854}$} & \cellcolor{green!20}{$\uparrow \mathbf{11.296}$} & \cellcolor{green!20}{$\uparrow \mathbf{4.538}$} & \cellcolor{green!20}{$\uparrow \mathbf{2.084}$}
& \cellcolor{green!20}{$\uparrow \underline{0.969}$} & \cellcolor{green!20}{$\uparrow 1.465$} & \cellcolor{green!20}{$\uparrow \mathbf{10.151}$} & \cellcolor{green!20}{$\uparrow \mathbf{2.847}$} \\
\textbf{LS}
& \cellcolor{green!20}{$\uparrow \mathbf{0.512}$} & \cellcolor{green!20}{$\uparrow \underline{1.610}$} & \cellcolor{green!20}{$\uparrow \underline{4.435}$} & \cellcolor{red!20}{$\downarrow 3.490$}
& \cellcolor{red!20}{$\downarrow 0.073$} & \cellcolor{red!20}{$\downarrow 7.195$} & \cellcolor{green!20}{$\uparrow 1.747$} & \cellcolor{red!20}{$\downarrow 5.280$}
& \cellcolor{red!20}{$\downarrow 0.854$} & \cellcolor{green!20}{$\uparrow 0.993$} & \cellcolor{red!20}{$\downarrow 3.258$} & \cellcolor{red!20}{$\downarrow 0.003$}
& \cellcolor{red!20}{$\downarrow 0.927$} & \cellcolor{red!20}{$\downarrow 32.713$} & \cellcolor{red!20}{$\downarrow 37.833$} & \cellcolor{red!20}{$\downarrow 6.120$} \\
\textbf{GT}
& \cellcolor{green!20}{$\uparrow \underline{0.439}$} & \cellcolor{green!20}{$\uparrow 1.570$} & \cellcolor{red!20}{$\downarrow 0.806$} & \cellcolor{red!20}{$\downarrow 3.670$}
& \cellcolor{red!20}{$\downarrow 0.146$} & \cellcolor{green!20}{$\uparrow \underline{8.963}$} & \cellcolor{green!20}{$\uparrow 0.780$} & \cellcolor{green!20}{$\uparrow \underline{4.750}$}
& \cellcolor{red!20}{$\downarrow 2.054$} & \cellcolor{green!20}{$\uparrow 0.164$} & \cellcolor{green!20}{$\uparrow 0.462$} & \cellcolor{red!20}{$\downarrow 0.440$}
& \cellcolor{red!20}{$\downarrow 0.219$} & \cellcolor{red!20}{$\downarrow 21.593$} & \cellcolor{red!20}{$\downarrow 23.183$} & \cellcolor{green!20}{$\uparrow \underline{0.710}$} \\
\textbf{DO}
& \cellcolor{green!20}{$\uparrow 0.439$} & \cellcolor{red!20}{$\downarrow 1.233$} & \cellcolor{green!20}{$\uparrow 2.285$} & \cellcolor{red!20}{$\downarrow 1.100$}
& \cellcolor{red!20}{$\downarrow 0.073$} & \cellcolor{red!20}{$\downarrow 0.409$} & \cellcolor{green!20}{$\uparrow 0.301$} & \cellcolor{red!20}{$\downarrow 3.440$}
& \cellcolor{red!20}{$\downarrow 0.654$} & \cellcolor{red!20}{$\downarrow 3.749$} & \cellcolor{green!20}{$\uparrow 0.790$} & \cellcolor{red!20}{$\downarrow 0.180$}
& \cellcolor{red!20}{$\downarrow 1.512$} & \cellcolor{red!20}{$\downarrow 28.134$} & \cellcolor{red!20}{$\downarrow 28.828$} & \cellcolor{green!20}{$\uparrow 0.270$} \\
\textbf{JPG}
& \cellcolor{red!20}{$\downarrow 0.219$} & \cellcolor{green!20}{$\uparrow 0.157$} & \cellcolor{red!20}{$\downarrow 2.285$} & \cellcolor{red!20}{$\downarrow 1.280$}
& \cellcolor{red!20}{$\downarrow 1.219$} & \cellcolor{red!20}{$\downarrow 4.415$} & \cellcolor{red!20}{$\downarrow 0.134$} & \cellcolor{red!20}{$\downarrow 6.790$}
& \cellcolor{green!20}{$\uparrow \mathbf{1.000}$} & \cellcolor{green!20}{$\uparrow 1.000$} & \cellcolor{red!20}{$\downarrow 1.226$} & \cellcolor{red!20}{$\downarrow 0.270$}
& \cellcolor{green!20}{$\uparrow 0.927$} & \cellcolor{green!20}{$\uparrow \underline{1.711}$} & \cellcolor{green!20}{$\uparrow 7.194$} & \cellcolor{green!20}{$\uparrow 0.180$} \\
\textbf{BD}
& \cellcolor{red!20}{$\downarrow 0.146$} & \cellcolor{green!20}{$\uparrow 0.640$} & \cellcolor{red!20}{$\downarrow 0.941$} & \cellcolor{green!20}{$\uparrow \underline{1.540}$}
& \cellcolor{red!20}{$\downarrow 0.073$} & \cellcolor{red!20}{$\downarrow 0.327$} & \cellcolor{green!20}{$\uparrow \underline{1.075}$} & \cellcolor{green!20}{$\uparrow 1.950$}
& \cellcolor{red!20}{$\downarrow 1.073$} & \cellcolor{red!20}{$\downarrow 1.509$} & \cellcolor{green!20}{$\uparrow \underline{0.866}$} & \cellcolor{red!20}{$\downarrow 0.710$}
& \cellcolor{green!20}{$\uparrow \mathbf{1.324}$} & \cellcolor{green!20}{$\uparrow \mathbf{3.673}$} & \cellcolor{green!20}{$\uparrow 8.000$} & \cellcolor{red!20}{$\downarrow 6.030$} \\
\textbf{MED}
& \cellcolor{red!20}{$\downarrow 0.951$} & \cellcolor{red!20}{$\downarrow 4.177$} & \cellcolor{red!20}{$\downarrow 9.274$} & \cellcolor{red!20}{$\downarrow 2.700$}
& \cellcolor{green!20}{$\uparrow \mathbf{0.342}$} & \cellcolor{red!20}{$\downarrow 9.403$} & \cellcolor{red!20}{$\downarrow 7.930$} & \cellcolor{red!20}{$\downarrow 5.550$}
& \cellcolor{red!20}{$\downarrow 0.634$} & \cellcolor{red!20}{$\downarrow 0.478$} & \cellcolor{red!20}{$\downarrow 2.183$} & \cellcolor{red!20}{$\downarrow 5.320$}
& \cellcolor{red!20}{$\downarrow 1.146$} & \cellcolor{red!20}{$\downarrow 3.359$} & \cellcolor{green!20}{$\uparrow 1.817$} & \cellcolor{red!20}{$\downarrow 5.320$} \\
\textbf{RRP}
& \cellcolor{red!20}{$\downarrow 1.975$} & \cellcolor{red!20}{$\downarrow 7.120$} & \cellcolor{red!20}{$\downarrow 11.290$} & \cellcolor{red!20}{$\downarrow 0.800$}
& \cellcolor{red!20}{$\downarrow 1.999$} & \cellcolor{red!20}{$\downarrow 11.774$} & \cellcolor{red!20}{$\downarrow 17.876$} & \cellcolor{red!20}{$\downarrow 0.440$}
& \cellcolor{green!20}{$\uparrow \underline{0.707}$} & \cellcolor{red!20}{$\downarrow 0.070$} & \cellcolor{red!20}{$\downarrow 2.989$} & \cellcolor{red!20}{$\downarrow 0.710$}
& \cellcolor{green!20}{$\uparrow 0.634$} & \cellcolor{red!20}{$\downarrow 4.258$} & \cellcolor{red!20}{$\downarrow 1.409$} & \cellcolor{red!20}{$\downarrow 0.090$} \\
\textbf{HEQ}
& \cellcolor{red!20}{$\downarrow 0.805$} & \cellcolor{red!20}{$\downarrow 1.315$} & \cellcolor{green!20}{$\uparrow 0.538$} & \cellcolor{red!20}{$\downarrow 0.090$}
& \cellcolor{green!20}{$\uparrow \underline{0.561}$} & \cellcolor{green!20}{$\uparrow \mathbf{3.271}$} & \cellcolor{green!20}{$\uparrow \underline{2.016}$} & \cellcolor{green!20}{$\uparrow \mathbf{1.045}$}
& \cellcolor{red!20}{$\downarrow 0.746$} & \cellcolor{red!20}{$\downarrow 0.308$} & \cellcolor{green!20}{$\uparrow \underline{2.925}$} & \cellcolor{red!20}{$\downarrow 0.090$}
& \cellcolor{red!20}{$\downarrow 0.073$} & \cellcolor{green!20}{$\uparrow 1.164$} & \cellcolor{green!20}{$\uparrow 1.253$} & \cellcolor{red!20}{$\downarrow 0.090$} \\
\textbf{RSE}
& \cellcolor{red!20}{$\downarrow 71.763$} & \cellcolor{red!20}{$\downarrow 25.681$} & \cellcolor{red!20}{$\downarrow 60.753$} & \cellcolor{green!20}{$\uparrow \mathbf{0.180}$}
& \cellcolor{red!20}{$\downarrow 0.024$} & \cellcolor{red!20}{$\downarrow 8.749$} & \cellcolor{red!20}{$\downarrow 6.452$} & \cellcolor{green!20}{$\uparrow \underline{3.460}$}
& \cellcolor{red!20}{$\downarrow 0.073$} & \cellcolor{green!20}{$\uparrow \underline{1.755}$} & \cellcolor{red!20}{$\downarrow 2.586$} & \cellcolor{red!20}{$\downarrow 0.350$}
& \cellcolor{red!20}{$\downarrow 1.146$} & \cellcolor{green!20}{$\uparrow 0.101$} & \cellcolor{green!20}{$\uparrow \underline{9.672}$} & \cellcolor{red!20}{$\downarrow 0.180$} \\
\bottomrule
\multicolumn{17}{l}{\footnotesize Per column, \textbf{bold} = largest improvement, \underline{underline} = second largest among improving (green) cells.}\\
\end{tabular}
}
\vspace{-15pt}
\label{tab:lisa-delta-compact}
\end{table*}

Conventional defenses show inconsistent and often counterproductive
effects across both datasets.
Train-time methods generally degrade adversarial robustness despite
being designed to improve generalization: on GTSRB, label smoothing
and Gaussian transformation reduce shadow accuracy by up to
$-18.71$\,pp and $-23.30$\,pp respectively for ViT-B/16, and on
LISA their effects collapse entirely for ViT-B/16
($-37.83$\,pp and $-23.18$\,pp on AE-light).
Preprocessing defenses improve RP2 robustness in isolated cases
(JPEG: $+8.74$\,pp ResNet-18 on GTSRB; bit-depth: $+9.03$\,pp
ViT-B/16 on GTSRB) but consistently hurt shadow accuracy and, in
the case of random resize-and-pad, severely degrade benign accuracy
by up to $-22.63$\,pp on GTSRB.
Randomized smoothing causes catastrophic benign degradation on LISA
for ResNet-18 ($-71.76$\,pp), indicating instability on that
distribution.
In contrast, LAMDA improves every evaluated adversarial split across all backbones and both datasets, while preserving or improving benign accuracy in seven of eight dataset–backbone settings.

Across both datasets, CNN backbones (ResNet-18, ResNet-34) show larger
absolute shadow gains on GTSRB owing to their lower shadow baselines
(61.4\% and 56.7\% vs.\ 71.1\% for Swin-T and 68.1\% for ViT-B/16),
while on LISA the transformer backbones show the largest AE-light
improvements and CNNs lead on RP2.
Across both datasets and all three attack types, no backbone is left
unimproved by LAMDA on any adversarial split, confirming that the
method generalises across both architecture families.

\section{Ablation Study}
\label{sec:ablation}

We conduct a systematic ablation over the two auxiliary loss weights
$\lambda$ (alignment, $\mathcal{L}_{\text{align}}$) and
$\mu$ (prototype, $\mathcal{L}_{\text{proto}}$),
evaluating all combinations from
$(\lambda, \mu) \in \{0,1,2\}^{2} \setminus \{(0,0)\}$ (eight configurations in total)
against the CE-only $(0,0)$ baseline.
All other hyperparameters are held fixed throughout.
We train and evaluate each configuration on both GTSRB and LISA
across four backbones (ResNet-18, ResNet-34, Swin-T, ViT-B/16),
yielding 72 trained models in total.
$\Delta$ accuracy relative to $(0,0)$ is reported across four splits
(Benign, AE-light, AE-shadow, RP2) in
Figures~\ref{fig:gtsrb-ablation-delta} and~\ref{fig:lisa-ablation-delta},
with absolute values for $(1,1)$ in
Tables~\ref{tab:gtsrb-delta-compact} and~\ref{tab:lisa-delta-compact}.

\noindent \textbf{Each loss component contributes independently.}
Comparing single-component configurations
$(\lambda{=}1, \mu{=}0)$ and $(\lambda{=}0, \mu{=}1)$
reveals that the two losses target different aspects of robustness.
On GTSRB, $(\lambda{=}0, \mu{=}1)$ yields stronger AE-shadow gains
than $(\lambda{=}1, \mu{=}0)$ across all backbones
($+4.85$ vs.\ $+2.99$\,pp for ResNet-18;
$+3.52$ vs.\ $+2.50$\,pp for ResNet-34),
while $(\lambda{=}1, \mu{=}0)$ provides more consistent improvement
across all four splits simultaneously.
This asymmetry is consistent across both datasets:
$\mathcal{L}_{\text{proto}}$ drives the largest single-split gains
while $\mathcal{L}_{\text{align}}$ contributes broader, more uniform
improvement.


\noindent \textbf{The two losses are complementary, not additive.}
The joint $(\lambda{=}1, \mu{=}1)$ configuration substantially
exceeds both single-component settings on every split and backbone.
On GTSRB ResNet-18, the AE-shadow gain at $(1,1)$ is $+12.50$\,pp —
$2.6\times$ the $(\lambda{=}0, \mu{=}1)$ gain of $+4.85$\,pp
and $4.2\times$ the $(\lambda{=}1, \mu{=}0)$ gain of $+2.99$\,pp —
and this super-additive effect holds across all four backbones and
both datasets without exception.
Across the ablation study, $(1,1)$ provides the strongest overall performance across the evaluated splits, backbones, and datasets.
On GTSRB, AE-shadow gains are
$+12.50$\,pp (ResNet-18, from a baseline of $61.38\%$),
$+9.28$\,pp (ResNet-34, baseline $56.66\%$),
$+8.23$\,pp (ViT-B/16, baseline $68.11\%$), and
$+3.32$\,pp (Swin-T, baseline $71.07\%$).
AE-light and RP2 also improve consistently
(RP2: $+4.35$\,pp ResNet-18, $+5.09$\,pp ResNet-34,
$+2.22$\,pp Swin-T, $+1.89$\,pp ViT-B/16).
Importantly, clean accuracy is not sacrificed:
benign gains at $(1,1)$ are $+3.98$\,pp (ResNet-18)
and $+3.91$\,pp (ResNet-34), with smaller but still positive gains
for the transformer backbones ($+1.01$\,pp ViT-B/16,
$+0.51$\,pp Swin-T).

\input{Tables/ablation-delta}

On LISA, the same $(1,1)$ configuration again provides the strongest overall performance, while the dominant improvement shifts to AE-light:
$+13.20$\,pp (ResNet-34, from a baseline of $42.27\%$)
and $+11.30$\,pp (Swin-T, baseline $40.79\%$).
AE-shadow and RP2 also improve across all backbones
(e.g., shadow: $+6.45$\,pp ResNet-34, $+4.70$\,pp ResNet-18;
RP2: $+7.89$\,pp ResNet-34, $+3.57$\,pp ResNet-18).
The strong overall performance of $(1,1)$ across two distinct datasets indicates that this weight setting transfers across the evaluated data distributions.

\noindent \textbf{Gains do not scale monotonically with weight magnitude.}
Moving from $(1,1)$ to $(2,2)$ or from $(1,0)$ to $(2,0)$ consistently
reduces performance, showing that increasing either weight beyond 1
is counterproductive.
On GTSRB ResNet-18, $(\lambda{=}2, \mu{=}2)$ drops benign accuracy
by $-1.90$\,pp and AE-light by $-1.61$\,pp relative to the
CE-only baseline.
This is not simply a case of high weights being uniformly bad, however:
$(\lambda{=}2, \mu{=}1)$ retains positive AE-shadow gains on GTSRB
($+5.12$\,pp ResNet-18, $+3.43$\,pp ResNet-34, $+5.21$\,pp ViT-B/16),
showing that a moderate prototype weight can partially compensate for
an overweighted alignment loss on shadow attacks specifically.
The critical failure mode occurs when $\mu{=}2$ is combined with
any non-zero $\lambda$: these configurations consistently degrade
AE-light and benign accuracy across all backbones, suggesting that
over-constraining the prototype space suppresses the representational
flexibility needed for other attack types.

\begin{figure}[htbp]
    \centering
    \setlength{\tabcolsep}{4pt}
    \begin{tabular}{cccc}
        \adjustbox{valign=b}{\includegraphics[width=0.28\columnwidth]{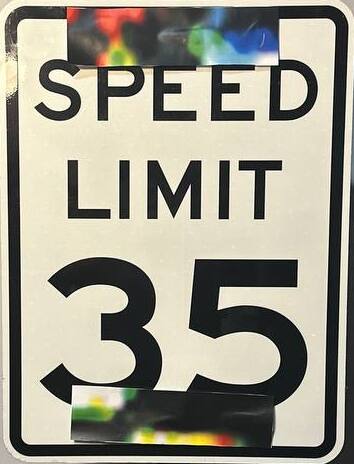}} &
        \adjustbox{valign=b}{\includegraphics[width=0.20\columnwidth]{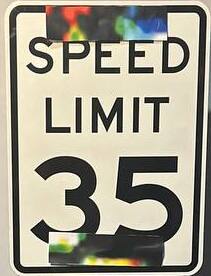}} &
        \adjustbox{valign=b}{\includegraphics[width=0.16\columnwidth]{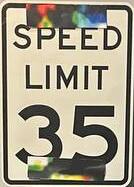}} &
        \adjustbox{valign=b}{\includegraphics[width=0.13\columnwidth]{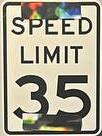}} \\[2pt]
        {\scriptsize 76\,cm} &
        {\scriptsize 203\,cm} &
        {\scriptsize 292\,cm} &
        {\scriptsize 458\,cm} \\
    \end{tabular}

    \vspace{4pt}
    \resizebox{\columnwidth}{!}{%
    \begin{tabular}{ll cccc}
        \toprule
        \textbf{Model} & \textbf{Setting} 
        & \textbf{76\,cm} & \textbf{203\,cm} & \textbf{292\,cm} & \textbf{458\,cm} \\
        \midrule
        \multirow{2}{*}{ResNet-18}
        & Baseline
        & \cmark~{\scriptsize SL35\,(0.38)} 
        & \xmark~{\scriptsize SigAh\,(0.40)} 
        & \xmark~{\scriptsize Sch\,(0.83)} 
        & \xmark~{\scriptsize Sch\,(0.72)} \\
        & $\lambda{=}1,\mu{=}1$
        & \cmark~{\scriptsize SL35\,(0.64)} 
        & \cmark~{\scriptsize SL35\,(0.48)} 
        & \cmark~{\scriptsize SL35\,(0.35)} 
        & \xmark~{\scriptsize TrnR\,(0.23)} \\
        \midrule
        \multirow{2}{*}{ResNet-34}
        & Baseline
        & \cmark~{\scriptsize SL35\,(0.42)} 
        & \xmark~{\scriptsize SigAh\,(0.45)} 
        & \xmark~{\scriptsize SigAh\,(0.95)} 
        & \xmark~{\scriptsize Sch\,(0.95)} \\
        & $\lambda{=}1,\mu{=}1$
        & \cmark~{\scriptsize SL35\,(0.47)} 
        & \cmark~{\scriptsize SL35\,(0.35)} 
        & \cmark~{\scriptsize SL35\,(0.33)} 
        & \cmark~{\scriptsize SL35\,(0.23)} \\
        \midrule
        \multirow{2}{*}{Swin-T}
        & Baseline
        & \cmark~{\scriptsize SL35\,(0.62)} 
        & \cmark~{\scriptsize SL35\,(0.41)} 
        & \xmark~{\scriptsize SigAh\,(0.37)} 
        & \xmark~{\scriptsize Sch\,(0.32)} \\
        & $\lambda{=}1,\mu{=}1$
        & \cmark~{\scriptsize SL35\,(0.59)} 
        & \cmark~{\scriptsize SL35\,(0.40)} 
        & \xmark~{\scriptsize Sch\,(0.35)} 
        & \xmark~{\scriptsize Sch\,(0.31)} \\
        \midrule
        \multirow{2}{*}{ViT-B/16}
        & Baseline
        & \cmark~{\scriptsize SL35\,(0.56)} 
        & \cmark~{\scriptsize SL35\,(0.50)} 
        & \xmark~{\scriptsize Mrg\,(0.29)} 
        & \xmark~{\scriptsize StpAh\,(0.31)} \\
        & $\lambda{=}1,\mu{=}1$
        & \cmark~{\scriptsize SL35\,(0.43)} 
        & \cmark~{\scriptsize SL35\,(0.46)} 
        & \cmark~{\scriptsize SL35\,(0.33)} 
        & \xmark~{\scriptsize Yld\,(0.27)} \\
        \bottomrule
    \end{tabular}%
    }

    \vspace{2pt}
    {\scriptsize \cmark~= correct (Speed Limit 35), \xmark~= misclassified (attack succeeded). Confidence in parentheses.}
    \caption{Physical adversarial patch attack (RP2) on a \emph{Speed Limit 35} sign evaluated at four distances. \textbf{Baseline}: standard training ($\lambda{=}0, \mu{=}0$). Our method ($\lambda{=}1, \mu{=}1$) improves robustness, correctly classifying the attacked sign at distances where the baseline fails.}
    \label{fig:physical_attack}
\end{figure}

\noindent \textbf{CNN and transformer backbones exhibit complementary sensitivity.}
On GTSRB, CNN backbones (ResNet-18, ResNet-34) show the largest absolute
AE-shadow gains at $(1,1)$, with ResNet-18 improving by $+12.50$\,pp
against a baseline of $61.38\%$.
Transformer backbones start from stronger baselines on shadow
(Swin-T: $71.07\%$, ViT-B/16: $68.11\%$) and correspondingly
show smaller absolute gains ($+3.32$\,pp and $+8.23$\,pp respectively).
On LISA the pattern inverts: transformers show the largest AE-light
improvements (Swin-T: $+11.30$\,pp, ViT-B/16: $+1.47$\,pp),
while CNNs dominate on RP2 and AE-shadow.
Across both datasets, no backbone is left unimproved by $(1,1)$ on any adversarial split, indicating consistent benefits across the evaluated architecture families.

\subsection{Language vs.\ Random Prototypes}
\label{sec:randbank}
Replacing the language banks with irrelevant-content banks of identical dimensionality and retraining with $(\lambda{=}1,\mu{=}1)$ on GTSRB ResNet-18 (CE-only baseline: $96.01\%$ benign, $61.38\%$ shadow) reduces shadow accuracy by $6.41$\,pp while benign drops by only $2.46$\,pp. 
In contrast, real descriptions yield gains of $+3.98$\ pp on benign inputs and $+12.50$\ pp under the shadow attack.
\section{Real-World Experiments}

We conduct a physical-world experiment using the RP2 adversarial patch attack~\cite{eykholt2018robust}. We generate printable adversarial patches targeting a \emph{Speed Limit 35} sign. The patches consist of two horizontal bars (covering approximately 19\% of the sign surface) printed on paper and physically attached to a real traffic sign. We then capture photographs of the attacked sign at four increasing distances: 76 cm, 203 cm, 292 cm, and 458 cm. Each image is classified by four backbone architectures (ResNet-18, ResNet-34, Swin-T, and ViT-B/16) under both the baseline and our proposed method ($\lambda{=}1, \mu{=}1$). 
As shown in Figure~\ref{fig:physical_attack}, the baseline models correctly classify the attacked sign in 6 out of 16 cases (37.5\%). In contrast, \textit{LAMDA} correctly classifies the sign in 12 out of 16 cases (75.0\%), corresponding to an improvement of 37.5 percentage points.


\section{Future work}
\label{sec:Future_works}

Physical attacks in our evaluation are generated against fixed target
models rather than the deployed network, an inherently transfer-based
setting. Extending the analysis to adaptive gradient-based attackers
with direct access to the backbone is left to future work.

\section{Conclusion}
\label{sec:conclusion}


We presented LAMDA, a training framework that improves the robustness of
traffic sign recognition models by anchoring visual representations to
frozen language prototypes derived from an off-the-shelf VLM. During
training, LAMDA uses two offline prototype banks and two complementary
auxiliary losses, requiring no adversarial examples and incurring no
inference-time overhead. Across GTSRB and LISA, four backbones, and three
physically realizable attacks, LAMDA is the only evaluated method that
consistently improves adversarial robustness across all attack--backbone--
dataset combinations, while preserving or improving clean accuracy in
nearly all cases. In particular, LAMDA achieves gains of up to
$+12.5$\,pp under shadow attacks on GTSRB and $+13.2$\,pp under
natural-light attacks on LISA, 
and improves physical RP2 robustness from $37.5\%$ to $75.0\%$ in our real-world experiment.
Ablations further show
that the alignment and prototype losses are complementary, with a single
$(\lambda{=}1,\mu{=}1)$ configuration generalizing across datasets and
architectures. These results suggest that language-grounded supervision
offers a practical path toward robust TSR without adversarial data,
deployment overhead, or substantial clean-accuracy tradeoffs.


\section*{Acknowledgments}
This work was supported in part by a grant from The BMW Group.

\bibliographystyle{IEEEtran} 
\bibliography{ref}

\end{document}